\documentclass[conference]{IEEEtran}
\IEEEoverridecommandlockouts
\usepackage{hyperref}
\usepackage{cite}
\usepackage{amsmath,amssymb,amsfonts}
\usepackage{algorithmic}
\usepackage{graphicx}
\usepackage{textcomp}
\usepackage{xcolor}
\def\BibTeX{{\rm B\kern-.05em{\sc i\kern-.025em b}\kern-.08em
    T\kern-.1667em\lower.7ex\hbox{E}\kern-.125emX}}

\begin{document}

\title{Benchmarking Catastrophic Forgetting Mitigation Methods in Federated Time Series Forecasting}

\author{
\IEEEauthorblockN{Khaled Hallak}
\IEEEauthorblockA{
\textit{CEA, List} \\
\textit{Université Paris-Saclay} \\
F-91120 Palaiseau, France \\
khaled.hallakhammoud@cea.fr}
\and
\IEEEauthorblockN{Oudom Kem}
\IEEEauthorblockA{
\textit{CEA, List} \\
\textit{Université Paris-Saclay} \\
F-91120 Palaiseau, France \\
oudom.kem@cea.fr}

}

\maketitle

\begin{abstract}
Catastrophic forgetting (CF) poses a persistent challenge in continual learning (CL), especially within federated learning (FL) environments characterized by non-i.i.d. time series data. While existing research has largely focused on classification tasks in vision domains, the regression-based forecasting setting prevalent in IoT and edge applications remains underexplored. In this paper, we present the first benchmarking framework tailored to investigate CF in federated continual time series forecasting. Using the Beijing Multi-site Air Quality dataset across 12 decentralized clients, we systematically evaluate several CF mitigation strategies, including Replay, Elastic Weight Consolidation, Learning without Forgetting, and Synaptic Intelligence. Key contributions include: (i) introducing a new benchmark for CF in time series FL, (ii) conducting a comprehensive comparative analysis of state-of-the-art methods, and (iii) releasing a reproducible open-source framework. This work provides essential tools and insights for advancing continual learning in federated time-series forecasting systems.
\end{abstract}

\begin{IEEEkeywords}
continual learning, federated learning, catastrophic forgetting, time series forecasting
\end{IEEEkeywords}

\section{Introduction}

In recent years, \emph{federated learning (FL)} has emerged as a powerful paradigm for distributed model training under strict data privacy constraints \cite{b1}, \cite{b2}. Rather than transmitting raw data, FL enables edge devices (clients) to train models locally and share only model updates with a central server, which aggregates them into a global model. This approach is well-suited for domains with sensitive, siloed, or geographically distributed data.

Among FL applications, \emph{Time Series Forecasting} remains relatively underexplored \cite{b3}, despite its importance in domains such as environmental monitoring, healthcare, and smart infrastructure. These systems often produce temporally-evolving, heterogeneous data across decentralized clients, making FL a natural fit. However, most FL research to date has focused on classification tasks\cite{b2}, neglecting the temporal and regression-specific challenges inherent to forecasting.

In dynamic real-world settings, forecasting models must continually adapt to seasonal variations, distributional shifts, and evolving task definitions. This need for continual adaptation motivates the integration of \emph{continual learning (CL)} \cite{b4}, \cite{b5}, which incrementally updates models as new data becomes available. However, such continual updates often induce \emph{catastrophic forgetting (CF)}—a phenomenon where the model forgets prior knowledge as it adapts to new tasks \cite{b6}, \cite{b10}.

In FL, this forgetting problem is exacerbated by non-i.i.d. data distributions across clients, asynchronous training schedules, and limited task overlap. When applied to time series forecasting, the challenge intensifies due to strong autocorrelations, lagged dependencies, and the regression-based nature of the tasks \cite{b3}, \cite{b8}. Notably, many widely-used CL algorithms—such as Elastic Weight Consolidation (EWC) \cite{b4}, Synaptic Intelligence (SI) \cite{b6}, Learning without Forgetting (LwF) \cite{b5}, and experience Replay \cite{b7}—have primarily been evaluated on classification benchmarks with i.i.d. data, offering limited insight into their applicability in temporally-structured federated settings.

This absence of systematic evaluation presents a critical gap, where existing frameworks either assume static objectives or ignore long-term retention under continual evolution and drift. To date, no prior work has proposed a rigorous benchmarking setup that targets CF in Federated Time Series Forecasting (FTSF).

\vspace{0.5em}
\noindent\textbf{Our Contributions:} To address this gap, we propose the first empirical framework for benchmarking CL methods in FTSF. Specifically:

\begin{itemize}
    \item We introduce a reproducible and extensible framework that integrates FL, CL, and multivariate regression, supporting lagged sequence generation, client-wise partitioning, and seasonal task scheduling. Codes supporting this work are publicly available online \footnote{\url{https://github.com/khaledhallakk/cf-federated-timeseries}}.
    \item We adapt and evaluate four core CL methods—Replay, EWC, SI, and LwF within a continual forecasting pipeline using LSTM models across decentralized clients.
    \item We employ fine-grained evaluation metrics, including task-wise forgetting, plasticity, and CPU computation time, to analyze the stability–plasticity trade-off.
    \item We validate the benchmark using real-world multivariate datasets exhibiting non-stationary temporal behavior and non-i.i.d. client distributions.
\end{itemize}

This work establishes a foundational benchmark for continual federated learning in time series domains, with a particular focus on mitigating CF. The benchmark encourages the development of robust and adaptive forecasting systems for real-world applications in IoT, edge-AI, and environmental monitoring.

This study is composed of five sections. First, we present the problem formulation, which integrates a review of related work on CL and CF in both federated and time series settings, identifies key limitations in the existing literature. Second, we detail the experimental setup, including dataset selection, task partitioning, preprocessing pipeline, learning protocols, evaluation metrics, and the CL methods under investigation. Third, we report and analyze the empirical results, focusing on performance, forgetting, and plasticity across forecasting tasks. Fourth, we provide a comprehensive discussion, contextualizing the findings and drawing insights into the stability–plasticity trade-off under Continual Federated Learning. Finally, we conclude with key insights, limitations, and future directions for advancing continual federated learning in real-world time series applications.

\section{Problem Formulation}

CL has emerged as a pivotal paradigm for machine learning in dynamic environments, where data evolves over time, and retraining from scratch is computationally prohibitive. Rather than learning from static, stationary datasets, CL enables models to adapt incrementally to new tasks or data distributions. This is particularly critical in real-world applications such as smart cities, healthcare, and industrial monitoring, where patterns change due to seasonality, policy shifts, or external events \cite{b12}, \cite{b13}. At the heart of CL lies the \emph{stability–plasticity dilemma}, which encapsulates the trade-off between acquiring new information and retaining previously learned knowledge. \emph{Plasticity} refers to the model’s capacity to rapidly learn and adapt to novel inputs, whereas \emph{stability} ensures the preservation of past knowledge and prevents CF. An effective CL system must strike a careful balance: too much plasticity causes the model to forget earlier tasks, while too much stability prevents it from adapting to new information. Failure to achieve this balance results in CF, a well-documented phenomenon wherein the model's performance on previous tasks degrades as it learns new ones \cite{b14}.

To address CF, several methods have been proposed: 
\begin{itemize}
    \item \textbf{Regularization-based approaches} such as EWC \cite{b4} and SI \cite{b6} by introducing constrains to important model parameters.
    \item \textbf{Replay-based strategies} retain a subset of prior data and replay it during training on new tasks\cite{b7}.
    \item \textbf{Knowledge Distillation-based methods}, including LwF \cite{b5}, transfer knowledge by mimicking previous model outputs.
\end{itemize}
Although effective for classification tasks (e.g., in MNIST or CIFAR-100), these methods are rarely evaluated under the constraints of temporal forecasting or decentralized learning.

\vspace{0.3em}
\noindent\textbf{Why Catastrophic Forgetting is Critical in Forecasting?} Time series forecasting introduces several unique challenges that exacerbate CF:
\begin{itemize}
    \item \textbf{Temporal autocorrelation}: Forecasting tasks are inherently ordered in time. Forgetting past regimes (e.g., winter pollution) degrades predictive reliability \cite{b15}.
    \item \textbf{Long-term dependencies}: In domains like medical, energy, and finance, historical patterns retain value across long horizons\cite{b16}.
    \item \textbf{Limited data access}: In federated learning (FL), raw data is decentralized, and client revisitation is rare due to privacy and connectivity constraints\cite{b1}.
    \item \textbf{Plasticity without memory}: High adaptability can lead to overfitting on recent data, harming model stability\cite{b13}.
\end{itemize}
For example, in a smart city scenario, an air quality sensor may experience recurring wintertime pollution surges. If a forecasting model forgets these patterns after training on summer data, the resulting predictions may fail to trigger timely interventions—leading to public health risks.

\vspace{0.3em}
\noindent\textbf{Gap in Existing Literature.} Although FL has gained traction for privacy-preserving collaboration across distributed clients \cite{b1}, its combination with CL—termed \emph{Federated Continual Learning (FCL)} has largely been limited to image classification settings \cite{b17}, \cite{b18}, \cite{b3}. Widely used benchmarks like Split-MNIST and class-incremental CIFAR-100 assume i.i.d. class distributions, fixed task boundaries, and synchronous updates \cite{b20}. These assumptions rarely hold in time series data, which features non-i.i.d. inputs, recurring seasonal patterns, and continuous drift.

In forecasting, time and causality are central: tasks are not interchangeable, and their ordering impacts both learning and forgetting. Yet, most CF studies in FL do not address this. Existing federated forecasting frameworks (e.g., Fed-Trend \cite{b21}, Fed-TIME\cite{b22}, Fed-PPBC \cite{b23}) either retrain periodically or neglect retention mechanisms entirely. As a result, there is no principled benchmark for studying CF in FTSF, despite its importance in edge-AI, IoT, and smart infrastructure.

\vspace{0.3em}
\noindent\textbf{Formal Problem Definition.} 

Let \(\mathcal{T}_1, \dots, \mathcal{T}_N\) denote a sequence of forecasting tasks ordered over time, where each task \(\mathcal{T}_i\) corresponds to a distinct temporal segment (e.g., a season). In a federated learning setting, we consider \(K\) clients. Each client \(k \in \{1, \dots, K\}\) holds a private multivariate time series dataset \(\mathcal{D}_k = \{\mathbf{x}^{(k)}_t, \mathbf{y}^{(k)}_t\}_{t=1}^{T_k}\), where \(\mathbf{x}^{(k)}_t \in \mathbb{R}^d\) represents the input features and \(\mathbf{y}^{(k)}_t \in \mathbb{R}\) the forecasting target at time \(t\). Due to privacy and storage constraints, clients are not permitted to retain or revisit earlier data once the corresponding task \(\mathcal{T}_i\) has been completed.

The goal is to learn a shared forecasting model \(f_\theta\), parameterized by \(\theta\), that minimizes the aggregated loss across all clients, while retaining performance on previously encountered tasks:

\[
\min_{\theta} \sum_{k=1}^K \frac{n_k}{n} \mathcal{L}_k(f_\theta; \mathcal{D}_k)
\]

s.t. performance on prior tasks $\mathcal{T}_i$ is retained, $\forall i < j$.
\begin{itemize}
  \item \(\mathcal{L}_k(f_\theta; \mathcal{D}_k)\) denotes the current task loss on client \(k\)'s local dataset \(\mathcal{D}_k\),
  \item \(n_k\) is the number of local samples at client \(k\), and \(n = \sum_{k=1}^K n_k\) is the total number of samples,
  \item \(f_{\theta_i}\) is the model trained immediately after completing task \(\mathcal{T}_i\),
  \item \(j\) is the index of the current task.
\end{itemize}

This setting introduces multiple challenges:
\begin{itemize}
    \item Tasks are temporally autocorrelated and non-i.i.d.
    \item Client data is heterogeneous and decentralized.
    \item Memory and computation constraints prevent global replay.
\end{itemize}

\vspace{0.3em}

\noindent\textbf{Contribution Overview.} 
The proposed benchmark is designed to address the challenges of continual forecasting under federated and non-i.i.d. conditions. Building on the above formulation, our framework enables the adaptation of core CL strategies to streaming, seasonal tasks over multivariate time series data in decentralized environments. By enforcing performance retention on prior tasks, the methodology supports rigorous evaluation of forgetting, plasticity, and learning dynamics across temporally evolving tasks.

\begin{figure}[htbp]
    \centering
    \includegraphics[width=\linewidth]{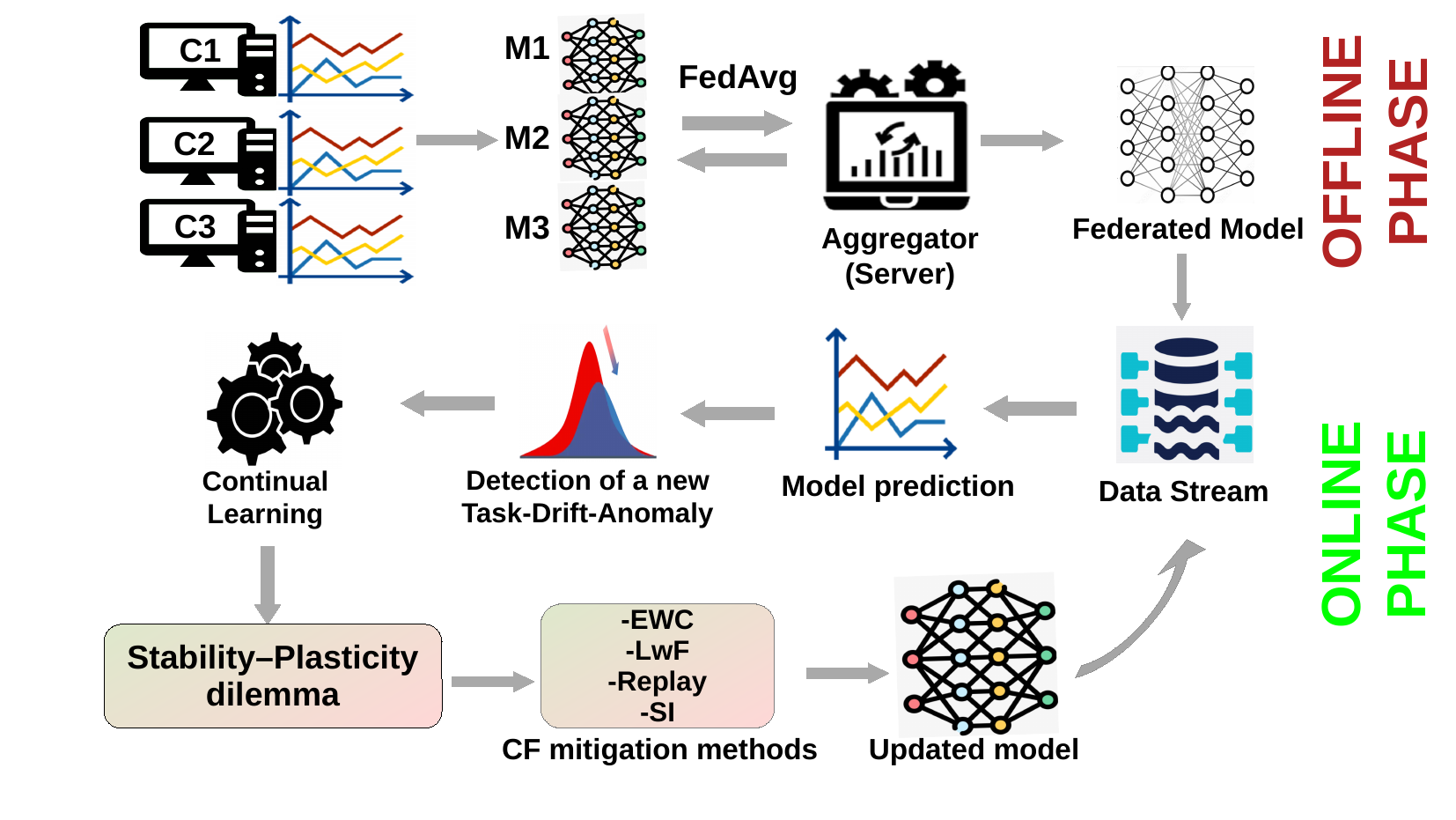}
    \caption{Overview of the FCL framework.}
    \label{fig3}
\end{figure}

Moreover, Figure~\ref{fig3} provides a high-level overview of the proposed FCL framework. The offline phase uses FedAvg to initialize a global forecasting model across clients. During the online phase, clients receive streaming data and monitor for task transitions, distributional drift, or anomalies. Upon detection of such events, appropriate CL mechanisms are applied locally to adapt the model while mitigating catastrophic forgetting. The framework addresses the stability–plasticity dilemma by integrating decentralized training with event-driven contthainual adaptation under memory and communication constraints.

\section{Experimental Setup}

\subsection{Dataset}
To evaluate CL strategies in realistic settings, we adopt the \textit{Beijing Multi-Site Air Quality} dataset—a well-established benchmark for spatiotemporal forecasting \cite{b24}. It provides hourly environmental measurements from multiple air quality monitoring stations across Beijing, covering the period from March 1, 2013, to February 28, 2017. Each station records multivariate features, including pollutant concentrations (PM2.5, PM10, NO2, CO, O3), meteorological factors (temperature, pressure, dew point, wind direction, wind speed), and temporal attributes (hour of day, day of week, season of the year).

To simulate decentralized learning, each station is treated as an independent client in the federated setting \cite{b1}, \cite{b8}. This naturally introduces non-i.i.d. local distributions due to spatial and environmental heterogeneity, thus emulating real-world constraints in edge learning systems \cite{b20}.

We focus on forecasting three representative target variables: \textbf{Temperature}, \textbf{PM2.5},  and \textbf{Wind Speed}. These targets exhibit distinct temporal characteristics: Temperature follows strong seasonal patterns with smooth daily cycles and high autocorrelation; PM2.5 concentrations are highly volatile and influenced by irregular pollution events, leading to weak seasonality and low autocorrelation; Wind Speed displays moderate seasonality and exhibits irregular short-term fluctuations driven by local atmospheric conditions. This diversity enables a comprehensive evaluation of CL strategies in terms of forgetting, plasticity, and long-term generalization across heterogeneous temporal dynamics.

\subsection{Continual Task Design}
To simulate non-stationary temporal dynamics within a FTSF context, we define each client's local data as a sequence of temporally ordered tasks. Formally, let $\mathcal{T}_1, \mathcal{T}_2, \ldots, \mathcal{T}_N$ denote a sequence of continual tasks, where each task $\mathcal{T}_i$ is a subset of the client’s time series representing a coherent temporal segment. These segments are defined such that:

- Each task $\mathcal{T}_i$ corresponds to a disjoint time window;

- The data in $\mathcal{T}_i$ becomes unavailable after training on task $\mathcal{T}_i$ is completed;

- Tasks are ordered chronologically to preserve the temporal structure inherent in forecasting problems \cite{b12,b18,b26}.

This design respects the fundamental CL assumption: data from previous tasks cannot be accessed unless explicitly stored (e.g., via a replay buffer) \cite{b7}, \cite{b27}. Furthermore, the temporal segmentation allows for the study of recurrent patterns and the impact of forgetting over long time horizons \cite{b2}, \cite{b28}.

To initialize the model with generalizable knowledge, we introduce a Base Task $\mathcal{T}_0$ composed of time series data that captures broad seasonal variability. This task is used in an offline phase where all clients collaboratively train a shared model using data from an extended, balanced period (e.g., one full year including all four seasons). The resulting model parameters serve as the starting point for all CL algorithms \cite{b4}, \cite{b6}.

In this work, we instantiate the above framework using the Beijing Multi-Site Air Quality dataset. The Base Task $\mathcal{T}_0$ comprises data from March 2013 to March 2014, covering all four meteorological seasons. Following this, we define $N = 11$ continual tasks ($\mathcal{T}_1, \ldots, \mathcal{T}_{11}$), each corresponding to one distinct season (Spring, Summer, Fall, Winter) over the period from 2014 to early 2017. This yields 11 disjoint tasks per client, chronologically ordered to simulate a realistic streaming scenario with seasonally evolving data. This setup enables the investigation of both short-term adaptation and long-term memory retention under FCL constraints \cite{b3}.

\subsection{Preprocessing Pipeline}
The data undergoes consistent preprocessing across clients, following best practices for time series forecasting:
\begin{itemize}
    \item \textbf{Missing values} are handled using forward and backward fill imputation;
    \item \textbf{Cyclical encoding} transforms periodic features (e.g., hour, day of week) via sine and cosine mappings;
    \item \textbf{Global normalization} uses robust min–max scaling based on the 1st and 99th percentiles, computed globally across all clients to preserve consistency;
    \item \textbf{Lagged sequences} are constructed by forming supervised windows: each input sample comprises $n$ past steps of multivariate features, predicting $p$ future steps of a selected target.
\end{itemize}
For each task, an 80\%-20\% temporal split is maintained between training and testing to preserve forecast realism.

\subsection{Learning Protocol}
We adopt a task-incremental CL protocol, executed in two stages:

\textbf{-Base Model Training:}

In the initial phase, all clients collaboratively train a global model using the \textbf{base task} data spanning the year 2013 to early 2014. Training is performed via the Federated Averaging (FedAvg) algorithm over 500 communication rounds, with each round consisting of 1 local epoch per round for each client. This extensive training allows the model—hereafter referred to as the BaseModel to capture generalizable spatiotemporal patterns. The resulting model parameters serve as the initialization point for all subsequent continual learning strategies.

\textbf{-Continual Learning:}
Following base model training, each client undergoes a sequence of $N = 11$ seasonal tasks, arranged in chronological order from 2014 to 2017. During each task, clients are exposed solely to their local data from the current season and train their models over 30 communication rounds, again using 1 local epoch per round. After local updates, clients send their model parameters to a central server, which performs aggregation using FedAvg. To mitigate CF, which naturally arises in sequential task learning, CL mechanisms are applied. The selected number of rounds and epochs was empirically determined to provide a stable trade-off between learning plasticity and stability, ensuring effective adaptation without overfitting or underlearning.

\subsection{Evaluation Metrics}
To evaluate both learning effectiveness and memory retention, we construct a task-wise performance matrix $P \in \mathbb{R}^{N \times N}$, where $P_{i,j}$ denotes the RMSE of model $M_i$ (trained up to task $\mathcal{T}_i$) on the test set of task $\mathcal{T}_j$. This matrix follows the continual learning protocol from \cite{b13}, \cite{b26}, allowing explicit measurement of stability and plasticity. From this matrix, we derive:

\begin{itemize}
    \item \textbf{Average Forgetting (AF):}
    \[
    \mathrm{AF} = \frac{1}{N-1} \sum_{j=1}^{N-1} (P_{N,j} - P_{j,j})
    \]
    Quantifies how much knowledge from earlier tasks is lost after completing all tasks. A lower AF indicates better stability across tasks (better memory retention).
    
    \item \textbf{Average Plasticity (AP):}
    \[
    \mathrm{AP} = \frac{1}{N} \sum_{j=1}^{N} P_{j,j}
    \]
    Reflects how effectively the model learns each new task, based on its performance on the current task’s test set right after training\cite{b13}. Lower values indicate better immediate learning.

    \item \textbf{Average Performance (AvgPerf):}
    \[
    \mathrm{AvgPerf} = \frac{1}{N} \sum_{j=1}^{N} P_{N,j}
    \]
    Indicates overall generalization ability of the final updated model across all previous tasks.
\end{itemize}

\subsection{Continual Learning Methods}
We benchmark six core CL strategies, all integrated into the same federated pipeline, consistent with recent CL taxonomies \cite{b13}, \cite{b26}, \cite{b28}.

\subsubsection{Naive Fine-Tuning (Original CL)}
Naive fine-tuning represents the simplest and most fundamental approach to CL. The model is sequentially trained on each new task using only its corresponding data, without any explicit mechanisms to preserve knowledge from previous tasks \cite{b13}, \cite{b22}.

\vspace{0.5em}
\noindent\textbf{Training Objective.}
Let $\theta$ denote the model parameters, and $(x_i, y_i)$ the current task's training samples. The loss function is simply the supervised objective:
\[
\mathcal{L}_{\text{naive}} = \frac{1}{N} \sum_{i=1}^{N} \| f_\theta(x_i) - y_i \|^2
\]

\vspace{0.5em}
\noindent\textbf{Characteristics.}
\begin{itemize}
    \item No access to data or models from previous tasks.
    \item Maximizes \textit{plasticity}, enabling fast adaptation to new data.
    \item Completely lacks mechanisms for \textit{stability}, making it highly susceptible to CF.
    \item Serves as a lower bound for memory retention.
    \item Provides a reference point to evaluate the effectiveness of more sophisticated CL strategies \cite{b13}.
\end{itemize}

\subsubsection{Replay Buffer (Experience Replay)}
Experience Replay mitigates CF by explicitly retaining and reusing a subset of past task data during training on new tasks \cite{b13}, \cite{b7}. By interleaving old and new samples, the model reinforces previously learned knowledge while learning new patterns.

\vspace{0.5em}
\noindent\textbf{Replay Construction.}
After completing a task, each client stores a small number of representative samples $(x_j^*, y_j^*)$ selected using a clustering-based strategy:
\[
x_j^* = \arg\min_{x \in \text{Cluster}_j} \| x - c_j \|_2
\]
where $c_j$ denotes the centroid of cluster $j$ obtained via $k$-means.

\vspace{0.5em}
\noindent\textbf{Training Objective.}
During training on a new task, both current-task data and replayed samples are used. The total loss becomes:
\[
\mathcal{L}_{\text{replay}} = \mathcal{L}_{\text{current}} + \lambda_{\text{replay}} \cdot \mathcal{L}_{\text{buffer}}
\]
where $\mathcal{L}_{\text{buffer}}$ is computed using samples from the replay buffer, and $\lambda_{\text{replay}}$ controls the replay strength.

\vspace{0.5em}
\noindent\textbf{Characteristics.}
\begin{itemize}
    \item Enhances stability by directly reinforcing prior knowledge.
    \item Requires memory storage but offers interpretable and sample-efficient mitigation \cite{b28}.
    \item Buffer size and sample selection strategy (e.g., clustering, importance sampling) greatly influence performance.
\end{itemize}

\subsubsection{Learning without Forgetting}
Learning without Forgetting (LwF) mitigates CF by preserving the functional output of previous models without storing past data \cite{b26}. It introduces a knowledge distillation loss that aligns the current model’s predictions with those of a frozen teacher model from the previous task.

\vspace{0.5em}
\noindent\textbf{Teacher–Student Paradigm.}
At the start of each new task:
\begin{itemize}
    \item A snapshot of the current model $\theta^*$ is saved as the \textit{teacher model}.
    \item The updated model $\theta$ (student) is trained to match the teacher’s output on current-task inputs.
\end{itemize}

\vspace{0.5em}
\noindent\textbf{Distillation Loss.}
Let $f_\theta(x)$ be the prediction of the current model and $f_{\theta^*}(x)$ the output of the frozen teacher. The knowledge distillation (KD) loss is defined as:
\[
\mathcal{L}_{\text{KD}} = \frac{1}{N} \sum_{i=1}^{N} \| f_\theta(x_i) - f_{\theta^*}(x_i) \|^2
\]

\vspace{0.5em}
\noindent\textbf{Total Loss.}
The training loss combines the new-task supervised loss $\mathcal{L}_{\text{new}}$ (e.g., MSE) and the distillation loss:
\[
\mathcal{L}_{\text{total}} = \mathcal{L}_{\text{new}} + \lambda_{\text{KD}} \cdot \mathcal{L}_{\text{KD}}
\]
where $\lambda_{\text{KD}}$ is a hyperparameter controlling the strength of output regularization.

\vspace{0.5em}
\noindent\textbf{Characteristics.}
\begin{itemize}
    \item LwF is data-efficient and avoids explicit replay.
    \item It is especially useful when replay buffers are impractical due to storage or privacy constraints.
    \item However, its effectiveness depends on how well the old and new task distributions align \cite{b26}.
\end{itemize}

\subsubsection{Elastic Weight Consolidation (EWC)}
Elastic Weight Consolidation (EWC) mitigates forgetting by selectively constraining updates to important parameters, based on the Fisher Information Matrix (FIM) \cite{b27}. It introduces a quadratic regularization term that penalizes deviation from parameters critical to prior tasks.

\vspace{0.5em}
\noindent\textbf{Regularization Loss.}
For each parameter $\theta_i$, the EWC penalty is given by:
\[
\mathcal{L}_{\text{EWC}} = \sum_i F_i (\theta_i - \theta_i^*)^2
\]
where:
\begin{itemize}
    \item $\theta_i^*$ is the parameter value after the previous task,
    \item $F_i$ is the estimated importance of $\theta_i$ from the FIM.
\end{itemize}

\vspace{0.5em}
\noindent\textbf{Total Loss.}
The full objective combines the new task loss with EWC regularization:
\[
\mathcal{L}_{\text{total}} = \mathcal{L}_{\text{new}} + \lambda_{\text{EWC}} \cdot \mathcal{L}_{\text{EWC}}
\]
where $\lambda_{\text{EWC}}$ controls the regularization strength.

\vspace{0.5em}
\noindent\textbf{Variants.}
\begin{itemize}
    \item \textbf{Classic EWC:} 
    \begin{itemize}
        \item $F_i$ is computed once after the base task.
        \item $\theta_i^*$ is fixed as the base model parameters.
        \item Static regularization throughout all tasks.
    \end{itemize}
    
    \item \textbf{Online EWC (O-EWC):} 
    \begin{itemize}
        \item $F_i$ and $\theta_i^*$ are updated after each task \cite{b28}.
        \item $F_i$ is accumulated using exponential decay:
        \[
        F_i^{(t)} = \gamma F_i^{(t-1)} + (1 - \gamma) \cdot \hat{F}_i
        \]
        where $\gamma \in [0, 1]$ is the decay rate and $\hat{F}_i$ is the new estimate.
        \item More flexible and adaptive to task dynamics.
    \end{itemize}
\end{itemize}

\vspace{0.5em}

EWC provides theoretical guarantees under sequential learning by anchoring parameters to their past optima \cite{b27}. The online variant improves scalability and adaptability in long task sequences.

\subsubsection{Synaptic Intelligence (SI)}
Synaptic Intelligence (SI) is a path-based regularization method that mitigates forgetting by continuously estimating the importance of each parameter during training \cite{b6}. Unlike EWC, which relies on Fisher information computed post-training, SI tracks how parameter updates contribute to the loss reduction during learning.

\vspace{0.5em}
\noindent\textbf{Importance Estimation.}  
During training on task $t$, the contribution of parameter $\theta_i$ is accumulated as:

\[
W_i \leftarrow W_i + \Delta \theta_i \cdot \left(-\frac{\partial \mathcal{L}}{\partial \theta_i}\right)
\]

\begin{itemize}
    \item $W_i$ is the accumulated contribution of parameter $\theta_i$ to the loss,
    \item $\Delta \theta_i$ is the change in $\theta_i$ during training,
    \item $\frac{\partial \mathcal{L}}{\partial \theta_i}$ is the gradient of the loss with respect to $\theta_i$.
\end{itemize}

Once the task concludes, the final importance score $\Omega_i$ is computed as:

\[
\Omega_i = \frac{W_i}{(\Delta \theta_i)^2 + \xi}
\]

where $\xi > 0$ is a small damping constant to ensure numerical stability.

\vspace{0.5em}
\noindent\textbf{Regularization Loss.}  
To protect important parameters, a quadratic penalty is applied to deviations from their previous values:

\[
\mathcal{L}_{\text{SI}} = \sum_i \Omega_i (\theta_i - \theta_i^*)^2
\]

where $\theta_i^*$ are the parameters at the end of the previous task.

\vspace{0.5em}
\noindent\textbf{Total Loss.}  
The final objective combines the standard regression loss with the SI regularization term:

\[
\mathcal{L}_{\text{total}} = \mathcal{L}_{\text{new}} + \lambda_{\text{SI}} \cdot \mathcal{L}_{\text{SI}}
\]

where $\lambda_{\text{SI}}$ weights the penalty on changes to important parameters from previous tasks.

\vspace{0.5em}

SI is efficient and adaptable, requiring no access to past data or external teachers \cite{b6}. It is robust to noise and suitable for CL scenarios with non-stationary, evolving data.

\section{Results and Discussion}

This section presents a comprehensive empirical evaluation of CL strategies within the context of FTSF. We benchmark performance over $11$ sequential tasks, each corresponding to a seasonal segment between 2014 and early 2017. The aim is to analyze the stability-plasticity trade-off across diverse temporal dynamics, using three forecasting targets of varying complexity. These targets were selected to capture distinct temporal structures. 

For all experiments, the input is constructed using a lagged sequence of $n=12$ hourly values, and the model predicts $p=6$ future time steps, aligned with common forecasting intervals in environmental monitoring systems.

Each CL method is evaluated using task-wise metrics: AF, AP, and AvgPerf. The values are scaled by $10^3$ to facilitate visual comparisons, without impacting the relative magnitudes or trends. All experiments are repeated over five independent trials with varying random seeds to ensure statistical robustness, and we report the mean $\pm$ standard deviation of each metric.

Hyperparameters were optimized using a grid search for each continual learning method and forecasting target to ensure fairness. The search aimed to resolve the stability–plasticity dilemma by achieving an AF below 0.1, thus minimizing forgetting while maintaining sufficient adaptability. Table~\ref{tab:hyperparams} summarizes the selected hyperparameters for each method across the three forecasting targets. This configuration provided optimal trade-offs for each case. For instance, LwF required increasing distillation weights ($\lambda_{\text{KD}}$) for harder targets, while SI also required increasing regularization weights ($\lambda_{\text{SI}}$) depending on the complexity of the forecasted target. Same for Replay methods showing target-specific sensitivity coefficient. Moreover, EWC and O-EWC performed best with a high regularization strength across all targets, depending on the obtained FIM. These settings were used consistently in all experiments and form the basis for the comparative analysis presented in the subsequent sections.

\begin{table}[htbp]
\caption{Optimized Hyperparameters per Method and Target}
\label{tab:hyperparams}
\centering
\begin{tabular}{lccc}
\hline
\textbf{Method} & \textbf{Temperature} & \textbf{PM2.5} & \textbf{Wind Speed} \\
\hline
$\lambda_{\text{KD}}$       & 30   & 60   & 120  \\
$\lambda_{\text{SI}}$        & 3  & 12  & 14 \\
$\lambda_{\text{Replay}}$ & 0.6 & 0.7 & 0.8 \\
$\lambda_{\text{O-EWC }}$ & $ 0.8\times 10^{6}$ & $ 1.5 \times 10^{6}$ & $ 1.3 \times 10^{6}$ \\
$\lambda_{\text{EWC}}$ & $ 10^{10}$  & $ 10^{6}$  & $ 10^{8}$\\
\hline
\end{tabular}
\end{table}

These parameter values reflect a balance between reducing CF and preserving adaptability across task transitions. The next sections interpret the results across CL methods and different forecasting targets, with further insights on method-specific trade-offs and computational efficiency.

\subsection{Importance of CL: Static Model vs Naive Adaptation}

In FTSF, adapting to evolving environmental dynamics is essential for maintaining long-term accuracy. To highlight the benefits of continual adaptation, we begin by comparing two foundational paradigms: (i) a static model trained only once on an initial dataset and applied to all future tasks without adaptation (\textbf{Static- No CL}), and (ii) a naive CL model that updates sequentially using only the current task's data, without applying any forgetting mitigation (\textbf{Naive-CL}).

\begin{figure}[htbp]
    \centering
    \includegraphics[width=\linewidth]{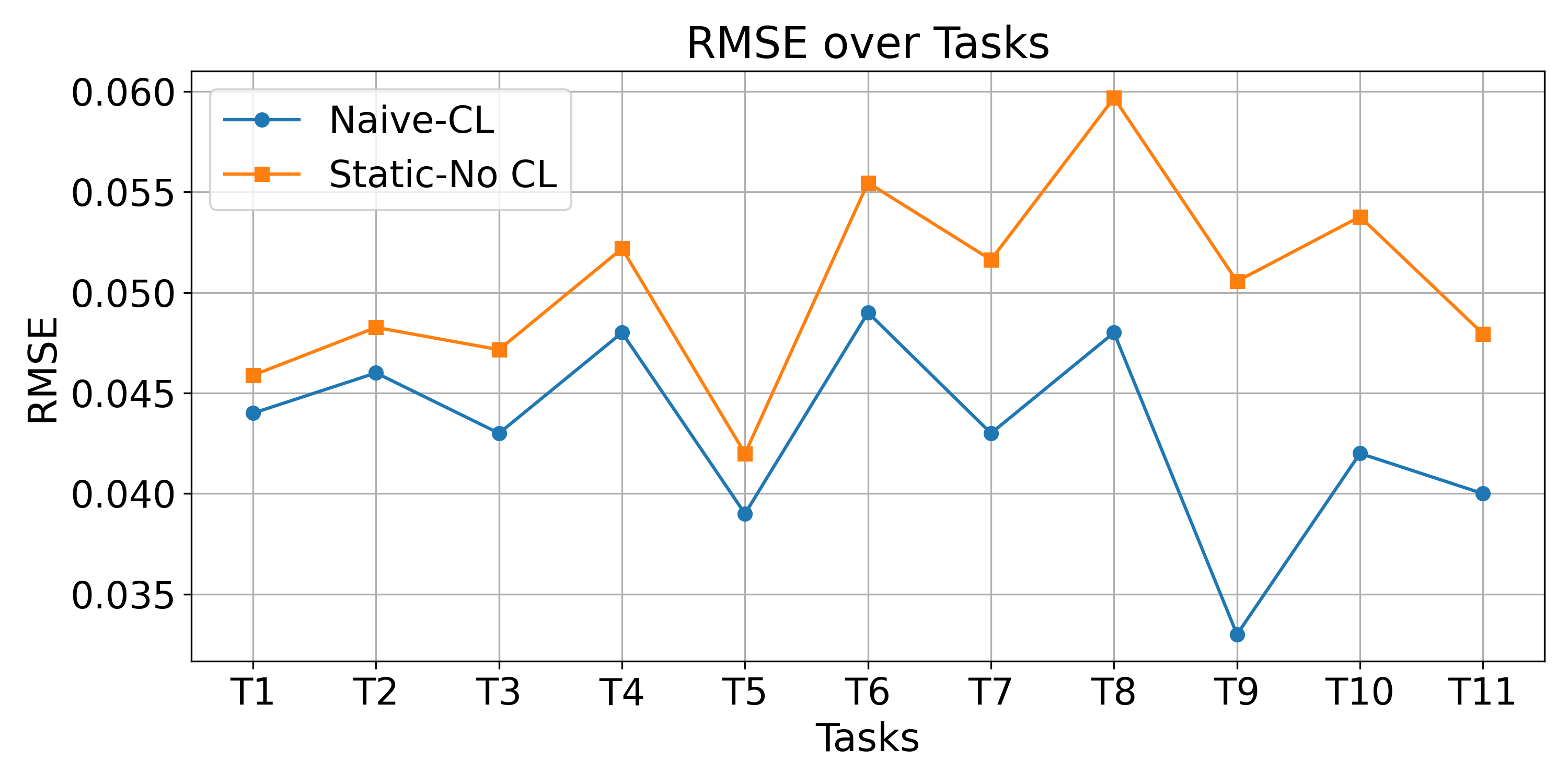}
    \caption{Comparison between the static BaseModel and Naive-CL on the Temperature forecasting target across continual tasks ($\mathcal{T}_1$ to $\mathcal{T}_{11})$}
    \label{fig1}
\end{figure}

\begin{table*}[htbp]
\centering
\caption{Comparison of CL methods across three target variables: Temperature, PM2.5, and Wind Speed. Each cell reports mean ± standard deviation over 5 trials. Best performances (lowest values for all metrics) are highlighted in bold.}

\resizebox{\textwidth}{!}{%
\begin{tabular}{|l|cccc|cccc|cccc|}
\hline
 & \multicolumn{4}{c|}{\textbf{Temperature}} & \multicolumn{4}{c|}{\textbf{PM2.5}} & \multicolumn{4}{c|}{\textbf{Wind Speed}} \\
\textbf{Method} & AF\(\downarrow\) & AP\(\downarrow\) & AvgPerf\(\downarrow\) & CPU\(\downarrow\) & AF\(\downarrow\) & AP\(\downarrow\) & AvgPerf\(\downarrow\) & CPU\(\downarrow\) & AF\(\downarrow\) & AP\(\downarrow\) & AvgPerf\(\downarrow\) & CPU\(\downarrow\) \\

\hline
NaiveCL & $1.32 \pm 0.06$ & $43.28 \pm 0.05$ & $44.94 \pm 0.02$ & $\textbf{442} \pm \textbf{34}$ & $5.25 \pm 0.10$ & $80.52 \pm 0.07$ & $81.99 \pm 0.08$ & $\textbf{437} \pm \textbf{30}$ & $8.44 \pm 0.23$ & $136.69 \pm 0.02$ & $141.75 \pm 0.25$ & $\textbf{428} \pm \textbf{32}$ \\

Replay & $-\textbf{0.63} \pm \textbf{0.05}$ & $\textbf{42.60} \pm \textbf{0.04}$ & $\textbf{42.35} \pm \textbf{0.03}$ & $673 \pm 40$ & $\textbf{-0.03} \pm \textbf{0.03}$ & $\textbf{79.19} \pm \textbf{0.06}$ & $\textbf{75.31} \pm \textbf{0.06}$ & $730 \pm 21$ & $\textbf{-0.16} \pm \textbf{0.05}$ & $\textbf{135.61} \pm \textbf{0.04}$ & $\textbf{132.21} \pm \textbf{0.04}$ & $690 \pm 20$ \\

KD & $-0.06 \pm 0.03$ & $43.62 \pm 0.01$ & $43.91 \pm 0.02$ & $515 \pm 29$ & $0.02 \pm 0.04$ & $80.43 \pm 0.02$ & $76.88 \pm 0.03$ & $538 \pm 30$ & $0.08 \pm 0.02$ & $136.74 \pm 0.02$ & $133.28 \pm 0.02$ & $552 \pm 48$ \\
O-EWC & $0.07 \pm 0.07$ & $44.21 \pm 0.03$ & $44.49 \pm 0.06$ & $554 \pm 36$ & $0.04 \pm 0.05$ & $79.44 \pm 0.03$ & $75.63 \pm 0.06$ & $581 \pm 19$ & $0.08 \pm 0.03$ & $137.03 \pm 0.02$ & $133.58 \pm 0.02$ & $559 \pm 44$ \\
EWC & $0.05 \pm 0.06$ & $50.46 \pm 0.05$ & $50.74 \pm 0.04$ & $521 \pm 18$ & $0.05 \pm 0.03$ & $79.79 \pm 0.05$ & $75.88 \pm 0.07$ & $571 \pm 45$ & $0.05 \pm 0.04$ & $137.82 \pm 0.01$ & $134.32 \pm 0.01$ & $542 \pm 25$ \\
SI & $0.01 \pm 0.03$ & $43.64 \pm 0.05$ & $43.93 \pm 0.02$ & $532 \pm 33$ & $0.02 \pm 0.03$ & $79.82 \pm 0.03$ & $76.03 \pm 0.02$ & $545 \pm 33$ & $0.08 \pm 0.02$ & $137.10 \pm 0.01$ & $133.75 \pm 0.02$ & $569 \pm 35$ \\
\hline
\end{tabular}%
}
\label{table1}
\end{table*}

As shown in Figure~\ref{fig1}, using the Temperature forecasting target as an illustrative example, Naive-CL consistently outperforms the static model across tasks. While the BaseModel captures general trends, its inability to adapt leads to growing prediction errors over time. In contrast, continual adaptation enables Naive-CL to capture season-specific and evolving patterns, improving accuracy even on later tasks. Notably, the performance gap between the static model and Naive-CL widens progressively as more tasks are introduced, highlighting the compounding benefits of CL in dynamic environment.

However, this plasticity comes at the cost of CF, a key drawback of naive CL. As the model adapts to new data, it overwrites parameters crucial to previous tasks, reducing its performance on earlier regimes. This is evident in the increasing AF values observed across the three forecasting targets (Table~\ref{table1}): 1.32 ,5.25 and 8.44 for  Temperature, PM2.5 and WindSpeed, respectively.

These results highlight the forgetting rate increases with the complexity and volatility of the target variable. While CL improves future-task prediction (as indicated by high AP: 43.28 for Temperature, 80.52 for PM2.5, and 136.69 for Wind Speed), the model's memory of earlier tasks deteriorates proportionally.

This motivates the need for CL strategies that balance plasticity (adaptation to new tasks) and stability (retention of old knowledge). In the following sections, we benchmark and analyze several CL methods.

\subsection{Evaluation of Core CL Algorithms: Forgetting vs Plasticity}

We now conduct a detailed evaluation of the five core CL strategies—Replay, EWC, O-EWC, LwF, and SI—in the context of FTSF. As reflected in Table~\ref{table1}, Replay emerges as the most consistently effective method across all three target variables. It achieves the lowest forgetting rates, often even negative (e.g., $-0.63$ for Temperature and $-0.16$ for Wind Speed), highlighting the occurrence of \emph{positive backward transfer}, where the acquisition of new knowledge enhances performance on prior tasks. Replay method also maintains strong plasticity and average performance, confirming its robustness. However, this effectiveness comes at a cost: Replay incurs the highest computational overhead, with runtime exceeding the next most expensive method by over 20\%.

In contrast, both EWC and O-EWC effectively reduced forgetting, achieving high stability across tasks. However, this came at the expense of plasticity, resulting in the lowest adaptability and overall performance among all methods. Notably, O-EWC outperformed standard EWC in plasticity due to its online update of the Fisher matrix, allowing more flexible regularization during task transitions.

LwF and SI offer a more balanced trade-off, showing slightly higher forgetting than Replay but preserving good plasticity, especially under dynamic conditions. Notably, LwF achieves strong adaptability due to its knowledge distillation mechanism, while SI leverages path-based parameter importance to enable smoother transitions across tasks. Both approaches benefit from careful hyperparameter calibration, which was instrumental in achieving favorable stability–plasticity trade-offs.

From a computational perspective, Naive CL is the most efficient due to its lack of auxiliary computations, while LwF and EWC maintain moderate CPU times, making them practical for constrained deployments. Replay methods, while highly effective, introduce overhead due to memory operations and auxiliary losses. SI and O-EWC are also computationally intensive, as they require frequent updates to importance matrices during training.

Overall, Replay demonstrates superior performance across all dimensions, but its reliance on stored data raises scalability and privacy concerns in federated settings. These results underscore the necessity for novel CL strategies tailored to time series forecasting that can achieve replay-level retention without storing any old data.

\subsection{Effect of Replay Sample Ratio on Plasticity}
To examine how the proportion of stored samples per task affects adaptability, we varied the percentage of task data retained in the buffer after each continual phase. Since average forgetting remained consistently near zero or slightly negative across all settings, we focus on how this \emph{replay sample ratio} influences plasticity.

\begin{figure}[htbp]
    \centering
    \includegraphics[width=\linewidth]{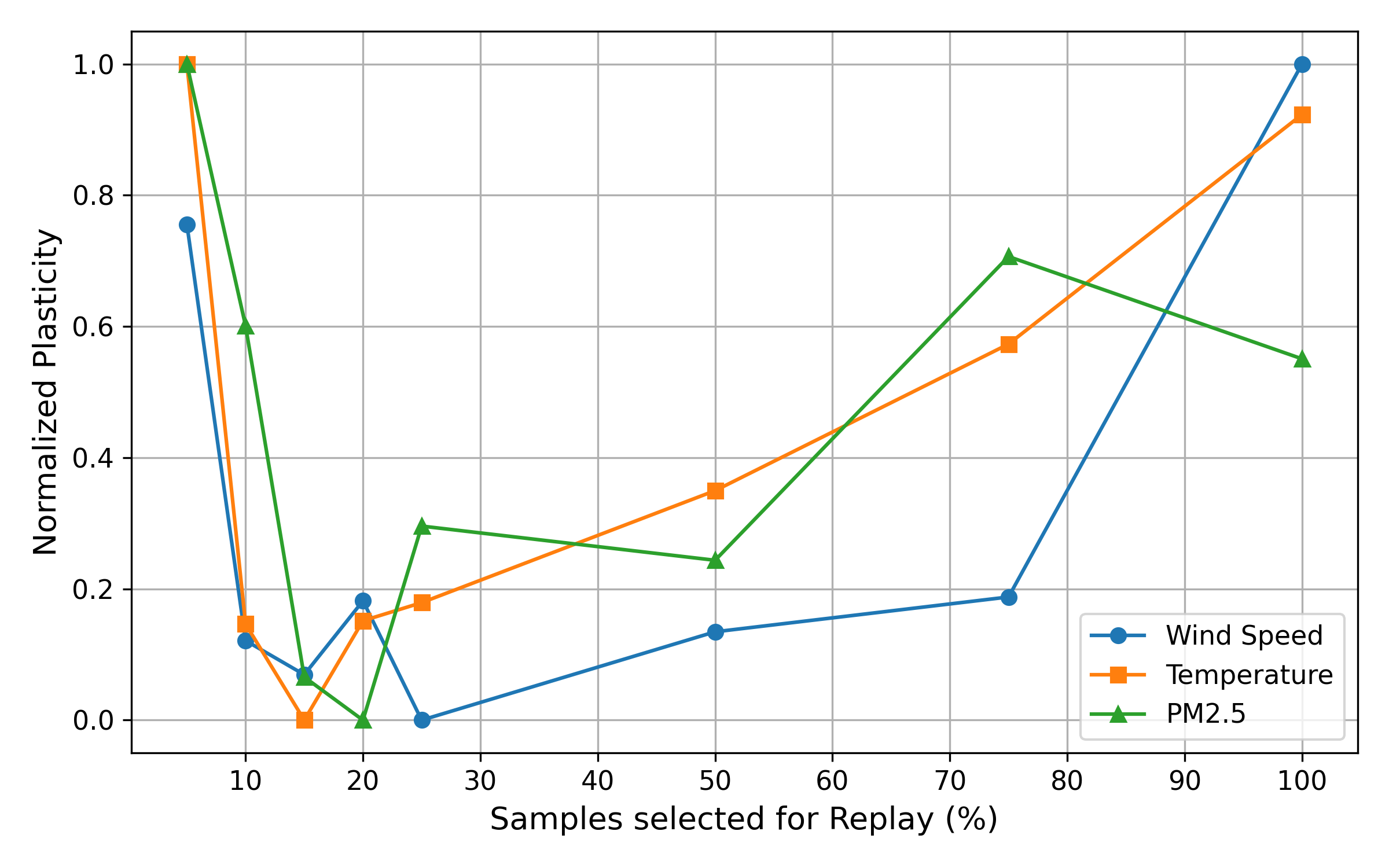}
    \caption{Effect of replay sample selection ratio on normalized plasticity}
    \label{buffer}
\end{figure}

Figure~\ref{buffer} shows the normalized plasticity scores across forecasting targets. Results indicate that the optimal ratio is target-specific: 15\% for \textit{Temperature}, 20\% for \textit{PM2.5}, and 25\% for \textit{Wind Speed}. Larger ratios led to degraded plasticity, likely due to increased redundancy and noise—even with representative sampling (e.g., $k$-means). This highlights the importance of tuning the replay ratio rather than maximizing it.

These findings highlight the importance of carefully tuning replay buffer size rather than maximizing it. While larger buffers may seem beneficial, they risk overwhelming the learner with outdated or overlapping information, thereby compromising the model’s responsiveness to new tasks. Future research could explore adaptive replay strategies where the buffer size is dynamically adjusted based on task difficulty, data drift, or information gain.

\section{Conclusion and Future Directions}
This study introduced a comprehensive benchmarking framework for evaluating CL algorithms in FTSF, a setting that remains under-explored despite its significance in real-world, distributed applications. By constructing a systematic evaluation protocol across multiple environmental forecasting targets—Temperature, PM2.5, and Wind Speed, we highlighted the varying difficulty levels of each prediction task and the unique challenges posed by temporal non-stationarity and cross-client heterogeneity. Our experiments revealed that replay-based strategies consistently achieve the best trade-off between forgetting mitigation and adaptive performance, albeit at the cost of increased memory, privacy, and efficiency concerns. Regularization-based methods such as EWC, Online EWC, SI, and LwF offer lightweight alternatives, but each suffers from specific limitations in stability or plasticity. Importantly, our analysis demonstrates the need for dynamic, task-aware CL methods tailored to the sequential structure of time series data. Moving forward, several research avenues emerge from this benchmark: the design of new CL algorithms that better preserve prior knowledge without sacrificing adaptability; the development of replay-free methods that retain efficacy under strict privacy constraints; the automation of hyperparameter selection through principled sensitivity analysis or meta-learning; and the extension of benchmarking protocols to handle longer prediction horizons, variable input lengths, and client-specific adaptations. Further, leveraging latent replay, prototype memories, or lightweight generative surrogates may offer promising directions for building memory-efficient models. Finally, future studies should explore explainable mechanisms to interpret forgetting dynamics, and establish open, standardized toolkits to accelerate reproducibility and innovation in this emerging research space. This study establishes a strong baselines and revealing critical limitations of current approaches, providing a foundation for advancing continual federated forecasting and fosters the development of robust, adaptive, and privacy-conscious learning systems.

\section{Acknowledgments}
This work was conducted in the framework of the AI-NRGY project (grant No: ANR-22-PETA-0004) funded by the PEPR
TASE program of France 2030.

\end{document}